%% file: main.tex
\documentclass[10pt,conference,compsocconf,letterpaper]{IEEEtran} 

\usepackage{multicol}
\usepackage{comment}
\usepackage[pdftex]{color, graphicx}
\usepackage{subfig}
\usepackage{amsmath}
\usepackage{amssymb}
\usepackage[vlined]{algorithm2e}
\usepackage{url}
\usepackage{xcolor}
\usepackage{float}
\usepackage[T1]{fontenc}
\usepackage[utf8]{inputenc}
\usepackage{soul}
\usepackage[normalem]{ulem}
\usepackage[space]{cite}

\newcommand{\fixme}[1]{{\color{red}\em\bf{[FIXME: #1]}}}

\newcommand{\shakshi}[1]{{\color{blue} *** {\textbf{Shakshi: }}\color{blue}{#1}}{\color{blue} ***}}

\newcommand{\PS}[1]{{\color{magenta} *** {\textbf{Shakshi: }}\color{magenta}{#1}}{\color{magenta} ***}}

\title{ 
Identifying Possible Rumor Spreaders on Twitter: A Weak Supervised Learning Approach 
}

\author{\IEEEauthorblockN{Shakshi Sharma and Rajesh Sharma}
\IEEEauthorblockA{\textit{Institute of Computer Science} \\
\textit{University of Tartu, Estonia}\\
Email: \{shakshi.sharma, rajesh.sharma\}@ut.ee}
}

\begin{document}
\maketitle

\begin{abstract}
Online Social Media (OSM) platforms such as Twitter, Facebook are extensively exploited by the users of these platforms for spreading the (mis)information to a large audience effortlessly at a rapid pace. 
It has been observed that the misinformation can cause panic, fear, and financial loss to society.
Thus, it is important to detect and control the misinformation in such platforms before it spreads to the masses.
In this work, we focus on rumors, which is one type of misinformation (other types are fake news, hoaxes, etc).
One way to control the spread of the rumors is by identifying
users who are possibly the rumor spreaders, that is, users who are often involved in spreading the rumors.
Due to the lack of availability of rumor spreaders labeled dataset (which is an expensive task), we use publicly available PHEME dataset, which contains rumor and non-rumor tweets information, and then apply a weak supervised learning approach to transform the PHEME dataset into rumor spreaders dataset. 
We utilize three types of features, that is, user, text, and ego-network features, before applying various supervised learning approaches.
In particular, to exploit the inherent network property in this dataset (user-user reply graph), we
explore Graph Convolutional Network (GCN), a type of Graph Neural Network (GNN) technique.
We compare GCN results with the other approaches: SVM, RF, and LSTM. Extensive experiments performed on the rumor spreaders dataset, where we achieve up to 0.864 value for F1-Score and 0.720 value for AUC-ROC, shows 
the effectiveness of our methodology for identifying possible rumor spreaders using the GCN technique.


\end{abstract}

\input{intro} %
\input{related} %
\input{Dataset}  %
\input{Eval}%
\input{conclusion} 

\bibliographystyle{IEEEtran}
\bibliography{main}
\label{sec:References}
\end{document}

%% file: intro.tex
\section{Introduction}\label{sec:intro}


Online  Social  Media  (OSM)  platforms,  initially  developed for  connecting  individuals have become a hotbed for users who spread misinformation  regularly  by  exploiting  the  connectivity  and globality  of  these  platforms \cite{kwak2010twitter}. On  one  side,  these  platforms  offer  a place  for  the  expression  of  views.  However,  on  the  flip  side, these  platforms  have  not  been  able  to  regulate  the  spread  of misinformation. According to \cite{mitnews}, false information propagates six times faster than the true news on these platforms, which can result in panic, fear, and financial loss to society \cite{zubiaga2018detection}. Thus, it is important to control misinformation propagation before it spreads to the masses. However, it is not a trivial task to identify such users  as  they  blend  very  well  by  creating  many  connections with users who are not involved in spreading misinformation activities \cite{fire2014friend}.



Misinformation can be categorized into two main types. The first one is the \textit{fake news}, the news which is certainly not true. The second one is the \textit{rumors}, a piece of information whose validity is in doubt at the time of posting. In other words, there is a doubt whether the information being posted is true or false. In this work, we focus on identifying the ``possible'' \textit{rumor spreaders}. We define \textit{rumor spreaders} as those users who are often engaged in spreading the rumors \cite{zubiaga2018detection}, and the term ``possible'' in our work points to the fact that it is very likely that the user could be a \textit{rumor spreader}.

In the past, researchers have proposed various techniques for identifying suspicious or malicious users involved in the spread of misinformation on OSM platforms such as Twitter. These works include analyzing the user profiles' information \cite{benevenuto2010detecting, thomas2013trafficking}, observing user activity patterns across a specific time window \cite{gupta2017towards, gurajala2015fake}, tracking the profiles through the usage of a smartphone's battery \cite{salehan2013social}. Furthermore, few additional 
approaches 
utilized graph-based techniques \cite{cao2012aiding} and multi-modal feature exploitation \cite{jiang2014detecting} for the detection of malicious profiles.
 
Identifying possible \textit{rumor spreaders} is important, as they could be the potential source of misinformation propagation. Curbing on such users means controlling the misinformation diffusion as well. However,
this problem has not acquired significant
attention in contrast to detecting \textit{rumors} or \textit{fake news}. This is primarily due to the lack of an annotated dataset about \textit{rumor spreaders} \cite{han2019neural}.
Thus, identifying possible \textit{rumor spreaders} is challenging in many aspects, which is the theme of this research work.

In this paper, we use the \textbf{PHEME} dataset, which
contains \textit{rumor} and \textit{non-rumor} tweets about five incidents, that is, i) \textit{Charlie hebdo}, ii) \textit{German wings crash}, iii) \textit{Ottawa shooting}, iv) \textit{Sydney siege}, and v) \textit{Ferguson} occurred between 2014 and 2015. 
In order to transform the tweets dataset into a rumor spreaders dataset\footnote{Code and data is available at - https://github.com/shakshi12/RumourGNN}, we explore the sentiments of the tweets and calculate the \textit{rumor spreaders' intensity score} (that is, how often a user posts \textit{rumor} tweets). In addition, we also utilize the weak supervised learning approach,
which is a branch of machine learning used to label the un-annotated data using few or noisy labels in order to avoid the expensive task of manual annotation of the data \cite{weaksupervision}. 
Please note that the labels that are generated using this approach are the \textit{near ground-truth} labels. Thus, we use the term `possible' for rumor spreaders dataset.


After the data transformation step, we leverage the following three distinct features for classifying possible \textit{rumor spreaders}:
\begin{enumerate}
\item \textbf{User Features: }This includes users' features such as \textit{number of followers, number of favorites}.

\item \textbf{Text Features: }We exploit users' tweets as a second set of features.

\item \textbf{Ego-Network Features:} 
We also explore the network of users who posted the tweets and the users who respond to those tweets. 
We illustrate this
using Figure \ref{fig:ego_network}, wherein the central node I is a node who has posted a tweet, which we refer to as initiator user and the nodes \textbf{R1, R2, R3}, and \textbf{R4} are the responder users who have replied to the user \textbf{I}'s tweet. The weights on the edges correspond to the number of times user \textbf{I}, and responder user interacted with each other. Based on our dataset, we are only able to create a one-hop network (ego-network) comprising of initiator user and its responders.
\begin{figure}[!htbp]
    \centering
    \includegraphics[width=8cm,height=8cm,trim = {6cm 6cm 8cm 7cm}, clip, keepaspectratio]{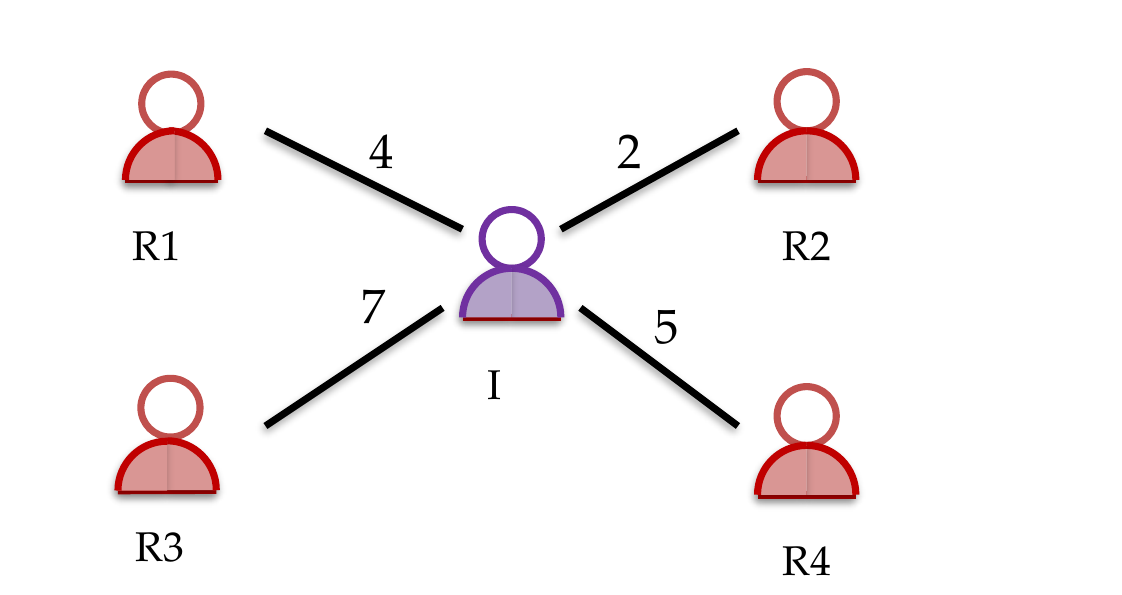}
    \caption{Undirected Weighted Ego-Network of a Twitter User }\label{fig:ego_network}
\end{figure}

\end{enumerate}

It is important to note that the network formed between the initiator of the tweets and their corresponding responders naturally calls for a technique that can exploit network features as well. Thus, in order to capture the network properties, we employ Graph Neural Network (GNN) based approach for identifying possible \textit{rumor spreaders}.
To be specific, we explore Graph Convolutional Network (GCN), a particular
type of GNN for our analysis.
In contrast to the baseline approaches: SVM, RF, and LSTM, the GCN approach
exhibit better performance and is able to achieve a value of 0.864 for F1-Score and 0.720 for AUC-ROC Score. 
To the best of our knowledge, this is the first work that has explored the \textbf{PHEME} dataset to identify possible \textit{rumor spreaders} using a weak supervised approach.



The rest of the paper is organized as follows.
Section \ref{sec:related} covers related work. Section \ref{sec:ds} covers dataset description and methodology used for identifying possible \textit{rumor spreaders}. Section \ref{sec:Evl} discusses experiments and their results, and Section \ref{sec:concl} concludes with some future directions.

%% file: related.tex
\section{Related work}\label{sec:related}

Researchers have proposed 
numerous techniques for identifying suspicious or fake user profiles across various OSM platforms such as Facebook \cite{gupta2017towards}, Twitter \cite{benevenuto2010detecting, wang2018sybilblind}, Tuenti  \cite{cao2012aiding}, to name a few. In this section, we discuss past studies across two dimensions. First, we 
discuss work related
to the identification of fake or suspicious users. Next, we 
present works about 
the detection of rumors. Our work lies at the intersection of these two types of works. 

\subsection{Suspicious Profile Detection}

Identification of suspicious (or fake) profiles on Twitter, a representative of OSM 
was initially 
analyzed using apparent features 
such as the \textit{number of  followers} and the \textit{number of followings} \cite{benevenuto2010detecting}, \textit{user-profile-name, screen-name,  and  email  parameters} \cite{thomas2013trafficking} 
Our work is different from these global approaches as they assumed the whole bird-eye view of the network. 
In contrast, we are only aware of the ego-networks. 
Some other works have focused on identifying suspicious followers by using multi-modal information \cite{jiang2014detecting}, which is out of the scope of this work.

It is a well-observed fact that mobile phones have been playing a 
vital role in the popularity of OSM platforms. 
Nonetheless, it has 
facilitated the rise of fake profiles 
as well \cite{salehan2013social}. 
Therefore, mobile phones too have
been used as a medium in tracking fake profiles. 
For instance, in \cite{perez2013dynamic}, the authors observed the daily behavior of users using mobile phone activity for detecting fake profiles. 
Besides researchers have 
also used camera-based sensors in detecting fake profiles within or across multi social networks \cite{bertini2015profile}.

\medskip

\subsection{Rumor Detection}

We assume that users who are often involved in disseminating rumors are more likely to be \textit{rumor spreaders} in comparison to users whose involvement in spreading rumor is less. Our approach for 
identifying the possible \textit{rumor spreaders} 
exploits (rumor and non-rumor) tweets 
along with other features present in the dataset. 
Therefore, we 
further present related literature with respect to rumor detection.


Initially starting with the theoretical framework for rumor spreading \cite{chierichetti2011rumor}, and later identification of rumors' temporal, structural, and linguistic features \cite{kwon2013prominent}, the topic related to rumor detection has attracted 
considerable attention in recent years, especially because of advancement in the field of artificial intelligence techniques. For example, in \cite{akhtar2018no}, 
various flavors of LSTM architecture are explored and in \cite{islam2019rumorsleuth}, a multi-task deep learning model is presented for rumor detection. In addition, 
several techniques such as the use of particle swarm optimization  \cite{kumar2019rumour}, multi-modal approach by exploiting textual as well as visual features from the data \cite{singhal2019spotfake, wang2018eann}, have 
also been 
examined in the literature.


Few approaches in rumor detection have also explored graph-based methodologies. 
Specifically, in \cite{rosenfeld2020kernel}, the authors constructed a Twitter follower graph and exploited diffusion patterns of
(mis) information for rumor detection. 
Some of the papers \cite{bian2020rumor, wei2019modeling, huang2019deep} have 
employed Graph Convolutional Network (GCN) based approaches for detecting rumors.  
Unlike these works, 
we study the rumors to 
identify possible \textit{rumor spreaders}.

User profiles have 
been analysed as an important aspect in detecting rumor propagation. Analysis of the user profile \cite{shu2019role}, identification of source of the rumor \cite{devi2020veracity}, genuineness score 
of the users in the social network which are spreading the rumors \cite{rath2017retweet} have been 
utilized in the past works. 
Our work lies at the boundary of these works as our aim is to 
identify possible \textit{rumor spreaders} by 
using not only 
the textual data that is being spread by these \textit{rumor spreaders} but also exploring their ego-networks. To accomplish this objective, we used GCN approach which has been mainly used in the past for 
identifying rumors and not for 
identifying possible \textit{rumor spreaders}. 

%% file: Dataset.tex
\section{Dataset Description and Methodology}\label{sec:ds}

In this section, we first discuss the \textbf{PHEME} dataset (Section \ref{sec:OrgDS}). Next, in Section \ref{sec:TranO2S}, we describe how we transform the \textbf{PHEME} tweets dataset into the rumor spreaders dataset. Finally, we explain three different types of features extracted from the dataset, which are provided as input to the machine learning algorithms (Section \ref{sec:FeatExt}).

\subsection{Original Dataset}\label{sec:OrgDS}

This paper utilizes the \textbf{PHEME}\footnote[1]{https://figshare.com/articles/PHEME\textunderscore dataset\textunderscore of\textunderscore rumours\textunderscore and\textunderscore non-rumours/4010619} dataset, which is a collection of \textit{rumor} and \textit{non-rumor} tweets that have been extensively used in previous works \cite{kochkina2018all, zubiaga2018detection}. The dataset comprises five events (or incidents) - \textit{Charlie hebdo}, \textit{German wings crash}, \textit{Ottawa shooting}, \textit{Sydney siege} and \textit{Ferguson}.
For the rest of this work, we refer to them as \textit{Charlie}, \textit{German}, \textit{Ottawa}, \textit{Sydney}, and \textit{Ferguson}, respectively.
The dataset contains information about the tweets pertaining to these incidents that have been posted as breaking news during the year 2014 -- 2015.
To be specific, data is provided in the form of five files, where each file is related to five particular incidents. The data in each file is stored in JSON format, having information about the source (or initiator's) tweet and its corresponding information.  Table \ref{Tbl:TweetDesdescp} provides detailed information about various fields. 

\begin{table}[!htbp]
\centering
\caption{Dataset Description}
\label{Tbl:TweetDesdescp}
\begin{tabular}{|l|l|l|}
\hline
\textbf{No} & \textbf{Fields}       & \textbf{Description}                               \\ \hline
\textbf{1}  & user id               & unique id of the initiator user                    \\ \hline
\textbf{2}  & tweet                 & tweet posted by initiator                          \\ \hline
\textbf{3}  & \# of followers       & of the initiator                                   \\ \hline
\textbf{4}           & \# of favorites       & of the initiator                                   \\ \hline
\textbf{5}           & verified user         & source user has verified account or not   \\ \hline
\textbf{6}           & reply user id         & unique id of the reply user                        \\ \hline
\textbf{7}           & reply tweet           & tweet posted by reply user                         \\ \hline
\textbf{8}           & \# of reply followers & of the reply user                                  \\ \hline
\textbf{9}           & \# of reply favorites & of the reply user                                  \\ \hline
\textbf{10}          & verified reply user   & reply user has verified account or not             \\ \hline
\textbf{11}          & label                 &  initiator's tweet is rumor or not \\ \hline
\end{tabular}
\end{table}

Furthermore, each source tweet has the ground-truth regarding
whether the tweet is a \textit{rumor} or \textit{non-rumor}. Table \ref{Tbl:DescpStat}, column `\# of Tweets (\%)' provides information about the number of \textit{rumor} and \textit{non-rumor} tweets for each of the incidents
for our analysis. 
\begin{table}[!htbp]
\centering
\caption{Distribution of Tweets and Spreaders for each of the incidents in the dataset}
\label{Tbl:DescpStat}
\begin{tabular}{|l|p{14mm}|p{14mm}|p{14mm}|p{14mm}|}
\hline
\textbf{} & \multicolumn{2}{|c|}{\textbf{\# of Tweets (\%)}} & \multicolumn{2}{|c|}{\textbf{\# of Spreaders (\%)}} \\ \hline
\textbf{Incidents} & \textbf{Rumor} & \textbf{Non-rumor} & \textbf{Rumor} & \textbf{Non-rumor} \\ \hline
\textbf{Charlie}        & 458 (22\%)                 & 1621 (78\%)             &      13879 (74.2\%)                       & 4821  (25.8\%)                \\ \hline
\textbf{German}   & 238 (50.7\%)               & 231 (49.3\%)                 & 1464  (50.3\%)                        & 1442  (49.7\%)                      \\ \hline
\textbf{Ottawa}      & 470 (52.8\%)               & 420 (47.2\%)                   & 3978  (51.1\%)                         & 3794  (48.9\%)                   \\ \hline
\textbf{Sydney}        & 522 (42.8\%)                 & 699 (57.2\%)               & 7545  (61.8\%)                        & 4658  (38.2\%)                   \\ \hline
\textbf{Ferguson}        & 284 (24.8\%)                 & 859 (75.2\%)                & 3792  (35\%)                        & 7001  (65\%)                   \\ \hline

\end{tabular}%
\end{table}

\subsection{Transformation of Tweets Dataset into Rumor Spreaders Dataset}\label{sec:TranO2S}

Due to the lack of an annotated dataset of users who are spreading the \textit{rumors} on OSM platforms, we first transform
the \textbf{PHEME} dataset (which carries information pertaining to the initiator's tweets, such as users who replied to the initiator's tweets, the \textit{followers count} of the initiator user, etc.) into the rumor spreaders dataset.
Table \ref{tbl:datadim}, column `Tweets', shows the original dimensions of the incidents, wherein each cell value represents the total number of initiator's tweets (36189 rows in case of \textit{Charlie}) and the total number of features, including ground-truth labels (11 columns for all the incidents).

\begin{table}[!htbp]
\caption{Dimensions of the \textbf{PHEME} dataset at various levels}
\label{tbl:datadim}
\begin{tabular}{|l|l|l|l|}
\hline
\textbf{Incidents}  & \textbf{Tweets} & \textbf{Rumor Spreaders} & \textbf{ Adjacency Matrix} \\ \hline
\textbf{Charlie}  & 36189, 11 & 18700, 304 &  18700, 18700 \\ \hline
\textbf{German}  & 4020, 11 & 2906, 304 &  2906, 2906 \\ \hline
\textbf{Ottawa}  & 11394, 11 & 7772, 304 &  7772, 7772 \\ \hline
\textbf{Sydney}  & 22775, 11 & 12203, 304 &  12203, 12203 \\ \hline
\textbf{Ferguson}  & 46064, 11 & 10793, 304 &  10793, 10793 \\ \hline
\end{tabular}
\end{table}

\begin{table*}[!htbp]
\caption{An Example of the Tweets on Ferguson incident}
\label{tbl:desribe}
\resizebox{\textwidth}{!}{%
\begin{tabular}{|l|l|l|l|}
\hline
\textbf{No} & \textbf{Initiator Tweet}                                                                                                                                                                      & \textbf{Reply Tweet 1}                                                                                                               & \textbf{Reply Tweet 2}                                                                                                               \\ \hline
1           & \begin{tabular}[c]{@{}l@{}}The mother of the boy killed in \#Ferguson\\ speaking to media about the loss of her son.\\ http://t.co/YlxEDKoebB\end{tabular}                                    & \begin{tabular}[c]{@{}l@{}}@AntonioFrench @b9AcE guess the cops were \\ protecting and serving the community again.\end{tabular}     & \begin{tabular}[c]{@{}l@{}}@AntonioFrench my heart aches \\ for her! This was so wrong!\end{tabular}                                 \\ \hline
2           & \begin{tabular}[c]{@{}l@{}}Police in \#Ferguson once charged a man \\ w/ destruction of property for bleeding \\ on their uniforms after they beat him \\ http://t.co/MRVP76sdUP\end{tabular} & \begin{tabular}[c]{@{}l@{}}@AnonyOps that's not true Dudeee!!\\ please go and read good newspapers. \\ \end{tabular} & \begin{tabular}[c]{@{}l@{}}@RianAlden not at all, but they \\ need to change some things at \\ \#ferguson PD. @AnonyOps\end{tabular} \\ \hline
\end{tabular}%
}
\end{table*}

In order to identify possible \textit{rumor spreaders}, we start by
placing each user with its corresponding tweets followed by its responders and their respective reply tweets. As part of the data cleaning process, we remove non-alphanumeric characters, URLs, stopwords, punctuations, lowercase all the words and perform additional Natural Language Processing operations as well, for instance, Porter Stemming of the words.
We cover the steps taken for the conversion:

\noindent\textbf{Step 1: Sentiment Analysis of the Reply tweets:}
It is highly likely that a tweet may have attracted multiple responses (or replies). To exemplify, Table \ref{tbl:desribe} shows an example of the two tweets from \textit{Ferguson} incident.
The first row corresponds to the \textit{non-rumor} tweet along with its replies (in this Table, we have shown only two replies, but a tweet can have any number of replies), whereas the second row corresponds to the \textit{rumor} tweet. 
It can be seen that the reply tweets possess sentiments with respect to the posted tweet, which can help in identifying the \textit{rumors}, and thus, \textit{rumor spreaders}. For instance,
\textit{non-rumor} tweets are in support of the initiator's tweet, hence, represents a positive sentiment.
Whereas in the case of \textit{rumor} tweets, reply tweets do not show support of the initiator's tweet, indicating negative sentiment. 
This could be a key indication that sentiments of the reply tweets play a key role in identifying whether the tweets posted by the user is \textit{rumor} or not.

\begin{figure}[!htbp]
    \centering
    \includegraphics[width=8cm,height=8cm,trim = {0cm 0cm 0cm 0cm}, clip,  keepaspectratio]{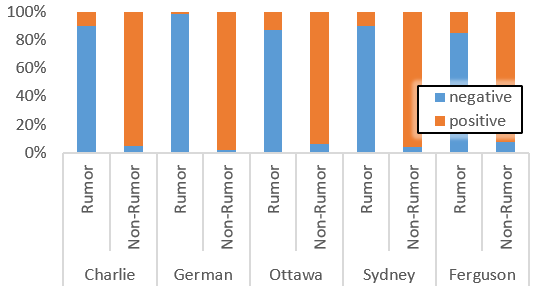}
    \caption{Sentiments of Reply Tweets with respect to Rumor and Non-Rumor Tweets for all the five incidents }\label{fig:Sentiments}
\end{figure}

In this regard, we first try to analyze the sentiments of the reply tweets using \textit{TextBlob API}\footnote[2] {https://textblob.readthedocs.io/en/dev/api\_reference.html}. Figure \ref{fig:Sentiments} displays the sentiments of the reply tweets with respect to \textit{rumor} and \textit{non-rumor} tweets for all five incidents. Specifically, the x-axis represents the positive and negative sentiments with respect to \textit{rumor} and \textit{non-rumor} reply tweets for all the incidents and y-axis corresponds to its percentage.
To capture the same, we consider the reply tweets under the \textit{rumor} category if the initiator's tweet is a \textit{rumor} otherwise \textit{non-rumor}. 
It is clear from Figure \ref{fig:Sentiments} that reply tweets under the \textit{rumor} category have mostly negative sentiments and vice-versa for all the incidents (we have excluded the \# of neutral sentiments, which are very few in number to avoid confusion). 
Thus, it can be validated that the sentiments of the reply tweets can be utilized to identify \textit{rumor} tweets, and hence, possible \textit{rumor spreaders}.

\noindent\textbf{Step 2: Labeling Reply tweets Using Weak Supervised Learning Approach:}
In order to identify the possible rumor spreaders, we would like to utilize the stance of reply tweets in our approach. However, no such information is present in the dataset. Thus, in  order  to  label  each  reply  tweet,
we apply the \textit{MinHash}\footnote[3] {snaPy API: https://pypi.org/project/snapy/} algorithm in line with \cite{nilizadeh2019think}. The \textit{MinHash} finds the similarities between each pair of the initiator's tweet and the reply tweet.
Specifically, if both these tweets are similar, then we assign the same label to the reply tweet as the initiator's tweet, indicating that the reply tweet is in support of the initiator's tweet. Otherwise, we assign the opposite label to the reply tweet. We considered two tweets to be similar if their similarity score is greater than or equal to 85\% (this threshold is validated manually). This approach of labeling the reply tweets is what we call as weak supervised learning approach due to the fact that we do not have the manual annotation of these tweets.

\begin{figure*}[ht]
    \centering
    \includegraphics[width=15cm,height=11cm,trim = {0cm 9cm 0cm 3cm}, clip, keepaspectratio]{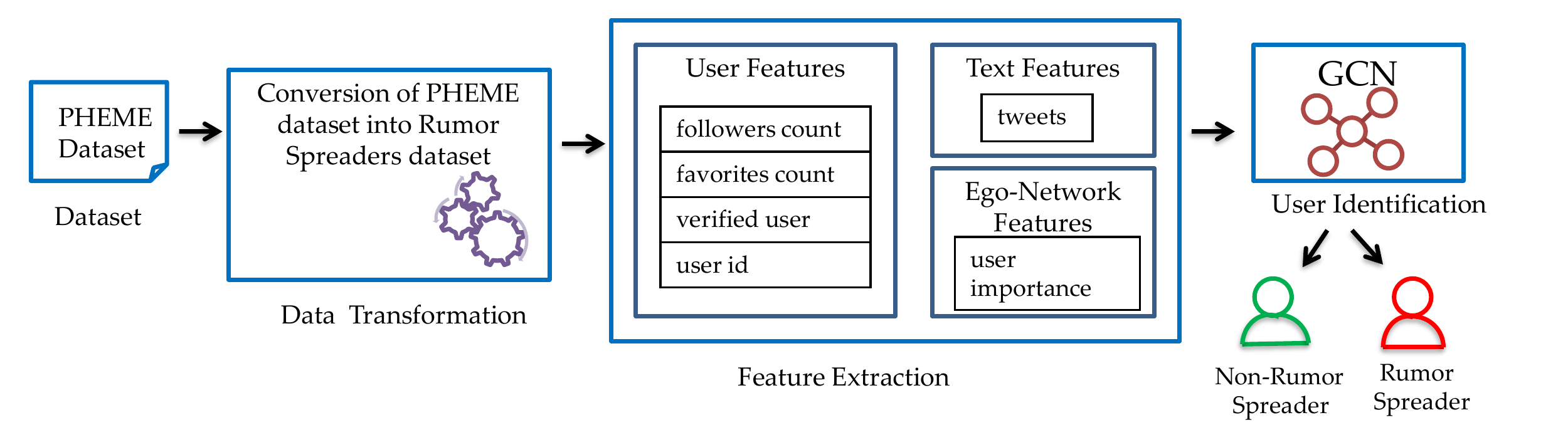}
    \caption{Framework}\label{fig:framework}
\end{figure*}

\noindent\textbf{Step 3: Calculating \textit{rumor spreaders' intensity score:}}
Tweets' labels only indicate whether a particular tweet is a \textit{rumor} or \textit{non-rumor}. Therefore, in order to identify possible \textit{rumor spreaders}, we calculate a score that indicates the intensity with which users spread \textit{rumors}, which we term as \textit{rumor spreaders' intensity score}.
We compute this score for each user by using the following formula:
    \begin{equation}
    score = \frac{\# \: of\: times\: user\: tweets\: rumor}{Total\: \#\: of\: times\: user\: tweets}
    \end{equation}

where the denominator is calculated by counting the total number of tweets posted by a user, whereas the numerator is calculated by counting the total number of \textit{rumor} tweets posted by a user.
The score range lies between [0, 1], where 0 means not a \textit{rumor spreader}, and 1, indicating possibly a \textit{rumor spreader}. 

In order to validate the effectiveness of this score, we calculate the degree (number of connections) from the user-user reply graph, as shown in Figure \ref{fig:ego_network}.
We observe that the nodes (or users) who are connected to many other nodes (or users), that is, high degree, are more involved in posting \textit{rumor} tweets as compared to nodes (or users) who have a low degree. We then manually check the users with their \textit{rumor spreaders' intensity score} calculated using Equation 1. The score is in line with the degree, which verifies our approach of identifying possible \textit{rumor spreaders}.


To model the problem as a binary classification problem, we put a threshold of $0.5$ to create two classes of users. That is, \textbf{\textit{if the \textit{rumor spreaders' intensity score} is < 0.5, then we assign 0 label (indicating non-rumor spreader class); otherwise, we assign 1 representing the possible rumor spreader class.}}
The reason for choosing this threshold is
based on the observation that the sentiments of the tweets as discussed in Step 1 are positive when the \textit{rumor spreaders' intensity score} is less than 0.5, which is indicative of \textit{non-rumor} spreaders class and vice-versa.
Based on this threshold conversion, Table \ref{Tbl:DescpStat}, column `\# of Spreaders (\%)'  shows the number of possible \textit{rumor spreaders} and \textit{non-rumor spreaders} for each of the five incidents in the dataset.

\subsection{Extraction of three sets of features}\label{sec:FeatExt}
In this section, we describe three distinct sets of features that we extract, to be utilized by our machine learning models for predicting possible \textit{rumor spreaders}.
\begin{enumerate}

    \item \textbf{User Features: }
    This
    set of features represents user's profile-based information that includes \textit{followers count}, \textit{favorites count}, and \textit{verified users}.
    
    \item \textbf{Text Features: }
    This feature represents text-related features, such as the \textit{tweet} column in our dataset. 
    We employ popular \textit{Word2Vec} embedding (a specific type of Vector Space Models) \cite{mikolov2013efficient} to convert the tweets into a numeric vector. Specifically, for each unique word in the \textit{tweet}, \textit{Word2Vec} generates its corresponding 300-dimension numeric vector, which is then aggregated in such a way that each sentence represents a 300-dimension vector. 
    Besides, we remove few noisy words as well, for instance, aaand, aand, aaaaand using English vocabulary.
    
    \item \textbf{Ego-Network Features}: 
  To 
  capture the network properties in the data, we 
  created a feature which we call as
  \textit{User Importance} (user\_imp) feature. This feature helps in adding the network properties in the dataset by calculating
  the importance of each user with respect to replies it has received. The formula for calculating this feature is as follows:  
    \begin{equation}
    user\_imp = \frac{\#\; of\; replies\; a\; user\; gets}{\#\; of\; replies\; of\; all\; users}
    \end{equation}

\end{enumerate}

Figure \ref{fig:framework} summarizes the framework, which we discussed in this Section. Section \ref{sec:Evl} discusses the parameters used by GCN and the results of various approaches.
    
\medskip

%% file: Eval.tex
\section{Experiments and Result Section}\label{sec:Evl}

In this section, we discuss few of the main hyper-parameter tuning used by our models (Section \ref{sec:parameter_tuning}). Next, we discuss various optimization techniques (Section \ref{sec:Metr}), and lastly, we discuss the results of our approaches (Section \ref{sec:Res}).

As mentioned in Section \ref{sec:intro}, there is a network formation between initiators and responders in terms of tweets. Thus, we use GCN \cite{kipf2016semi}, a type of GNN technique, which can exploit both the network structure and features (such as \textit{tweets, followers count}). In particular, we perform a binary classification for identifying if a specific Twitter user is possibly involved in \textit{rumor} spreading or not. In addition, we compare GCN with other baseline approaches (SVM, RF, and LSTM).

\subsection{Hyper-Parameters tuning}\label{sec:parameter_tuning}

Table \ref{tbl:parameter-tuning} shows the values for
various hyper-parameter settings that are chosen for fine-tuning the models for improving the performance. Here, we have specified few of the main hyper-parameters only.
For instance, 
SVM has a hyper-parameter, \textit{kernel},
which is a function that finds the similarity score between two data points even from the high dimensional input space in order to find the optimal hyperplane.
RF has a hyper-parameter known as, \textit{number of trees in forest} that specifies how many trees should be formed so that they can be used as a parallel estimators in order to make final prediction.
The rest of the hyper-parameters are specific 
to the neural networks (for both LSTM and GCN). The \textit{number of layers in a network} represents the total number of layers used in a model, \textit{number of channels in each layer} represents the total number of output neurons to be used in a layer, \textit{drop out layer} 
is used 
to
avoid overfitting of the data. The \textit{activation function} decides which neuron to activate for the next layer in the network, \textit{number of epochs} refers to the number of iterations the neural network model is to be trained for. The \textit{loss function} is used for refining the model after every epoch.
The \textit{NA} values in the table refer to Not Applicable.
    
\begin{table}[!htbp]
\centering
\caption{Hyper-Parameters Tuning of models}
\label{tbl:parameter-tuning}
\begin{tabular}{|l|l|l|l|l|}
\hline
\textbf{\begin{tabular}[c]{@{}l@{}}Parameters\textbackslash\\ Models\end{tabular}} & \textbf{SVM}                                                     & 
\textbf{RF}                                                     & 
\textbf{LSTM}                                                   & \textbf{GCN}                                                   \\ \hline
\textbf{Kernel}                                                                         & \begin{tabular}[c]{@{}l@{}}radial basis \\ function\end{tabular} & NA        & NA                                                      & NA                                                             \\ \hline
\textbf{\begin{tabular}[c]{@{}l@{}}\# of trees \end{tabular}} & NA     & 100                                                            & 2                                                               & 2                                                              \\ \hline

\textbf{\# of layers}                                                                         & \begin{tabular}[c]{@{}l@{}}NA \end{tabular} & NA & 2                                                             & 2
\\ \hline
\textbf{\begin{tabular}[c]{@{}l@{}}\# of channels\\ in first layer\end{tabular}} & NA           & NA                                                               & 32                                                              & 32                                                             \\ \hline

\textbf{\begin{tabular}[c]{@{}l@{}}\# of channels \\ in second layer\end{tabular}}      & NA    & NA                                                           & 2                                                               & 2                                                              \\ \hline
\textbf{Drop out layer}                                         & NA                        & NA                                                               & 0                                                              & 2                                                              \\ \hline
\textbf{\# of Epochs}                                                                      & NA      & NA                                                         & 300                                                             & 300                                                            \\ \hline
\textbf{\begin{tabular}[c]{@{}l@{}}Activation \\ function\end{tabular}}                 & NA        & NA                                                       & sigmoid                                                         & sigmoid                                                        \\ \hline
\textbf{Loss function}                                                                  & average & hinge                                                            & \begin{tabular}[c]{@{}l@{}}binary cross \\ entropy\end{tabular} & \begin{tabular}[c]{@{}l@{}}binary cross\\ entropy\end{tabular} \\ \hline
\end{tabular}
\end{table}
   
    Apart from the Table \ref{tbl:parameter-tuning} hyper-parameters, GCN model takes two additional inputs - graph adjacency matrix and nodes' features matrix. 
    The graph adjacency matrix stores the nodes' neighbors information in a $Z$X$Z$ matrix, where $Z$
    represents the total number of nodes (users) in the dataset. The nodes' features matrix $Z$X$F$ is the final matrix of the preprocessing step, where $Z$ represents the number of nodes and $F$ is the size of total features.
    As already mentioned, Table \ref{tbl:datadim}, column `Rumor Spreaders', shows the dimensions of the nodes' features matrix for each of the incident. 
    Specifically, each cell
    values represents the nodes, $Z$ and the features, $F$ where $F$ represents the features (attributes) such as \textit{followers count}, word embedding of tweet, \textit{user\_imp}. 
    In addition, the dimensions of the adjacency matrix are shown in \ref{tbl:datadim}, column `Adjacency Matrix',
    wherein each cell values represents the total number of users in a dataset.
    After providing required inputs to GCN, the model is trained to predict possible \textit{rumor} and \textit{non-rumor spreaders}.
   
  
    In order to predict possible \textit{rumor spreaders}, the nodes have labels as 0
    (\textit{non-rumor spreader}) or 1 (possibly a \textit{rumor spreader}). 
    In Section \ref{sec:Res}, we discuss results of all the machine learning approaches.

\subsection{Optimization Techniques}\label{sec:Metr}

As part of the optimization, we perform following steps -

\begin{enumerate}
    \item \textbf{Cross Validation:} To ensure the effectiveness of our model and to avoid overfitting of data, we perform K-Fold cross-validation on our dataset where \textit{K = 5}. To avoid the class imbalance problem in \textit{Charlie} incident, we use Stratified K-Fold.

    \item \textbf{Standardization: }All the features are standardized before training the machine learning model.
    \item \textbf{Feature Importance: }In addition to the above two techniques, three feature selection techniques, namely, \textit{Chi-Square, Information Gain, Gain Ratio} are applied to each of the five incidents of rumor spreaders dataset to check whether each feature is correlated with the target variable. 
    Table \ref{tbl:feature} depicts the p-values of three feature selection techniques. For all the features, p-value < 0.05 which shows that these features are important in predicting possible \textit{rumor spreaders}. Thus, we consider all the features in our experiments.
\end{enumerate}

\begin{table}[!htbp]
\caption{Feature Selection Techniques (values in the cells indicates their corresponding p-values)}
\label{tbl:feature}
\begin{tabular}{|l|l|l|l|}
\hline
\textbf{Features}& \textbf{Chi-Square} & \textbf{Information Gain} & \textbf{Gain Ratio} \\ \hline
\textbf{followers count}                           & 5.28e-05            & 4.62e-07                  & 8.56e-09            \\ \hline
\textbf{favorites count}                           & 8.66e-17            & 11.65e-22                 & 5.56e-11            \\ \hline
\textbf{verified users}                            & 2.26e-10            & 1.44e-05                  & 3.67e-07            \\ \hline
\textbf{user importance}                           & 7.87e-20            & 9.93e-23                  & 7.83e-26            \\ \hline
\end{tabular}
\end{table}

\begin{table}[!htbp]
\centering
\caption{Metrics performance of different models}
\label{Tbl:gw}
\begin{tabular}{|l|l|l|l|l|l|}
\hline
\textbf{S.No.} & \textbf{Metrics} & \textbf{SVM} & \textbf{RF} & \textbf{LSTM} & \textbf{GCN} \\ \hline
\multicolumn{6}{|c|}{\textbf{\textcolor{blue}{Charlie}}} \\ \hline
1 & \textbf{Accuracy}          &  0.748             & 0.760 &      0.671        &    \textbf{0.790}         \\ \hline
2 & \textbf{Precision}          &    0.748          & 0.758 &   0.571            &     \textbf{0.790}         \\ \hline
3 & \textbf{Recall}             &   0.748           & 0.628 &          0.571     &       \textbf{0.790}       \\ \hline
4 & \textbf{F1-Score}        &     0.853        & 0.840  &         0.778      &       \textbf{0.864}       \\ \hline
5 & \textbf{AUC-ROC}     &     0.600        & 0.600 &       0.570        &        \textbf{0.690}      \\ \hline
\multicolumn{6}{|c|}{\textbf{\textcolor{blue}{German}}} \\ \hline
6 & \textbf{Accuracy}           & 0.552             & 0.567 &       0.541        &    \textbf{0.715}          \\ \hline
7 & \textbf{Precision}          &  0.553 & 0.567            &         0.541      &      \textbf{0.717}        \\ \hline
8 & \textbf{Recall}             & 0.552        & 0.567     &         0.541      &      \textbf{0.716}        \\ \hline
9 & \textbf{F1-Score}         &   0.572           & 0.567 &        0.546       &       \textbf{0.709}       \\ \hline
10 & \textbf{AUC-ROC}      & 0.552             & 0.566 &         0.540      &       \textbf{0.720}       \\ \hline
\multicolumn{6}{|c|}{\textbf{\textcolor{blue}{Ottawa}}} \\ \hline
11 & \textbf{Accuracy}           &     0.567         & 0.565 &     0.552          &      \textbf{0.675}        \\ \hline
12 & \textbf{Precision}          & 0.567             & 0.565 &      0.552         &      \textbf{0.681}        \\ \hline
13 & \textbf{Recall}             &    0.566          & 0.565 &     0.552          &     \textbf{0.677}         \\ \hline
14 & \textbf{F1-Score}         &      0.578  & 0.569      &        0.559       &       \textbf{0.655}       \\ \hline
15 & \textbf{AUC-ROC}      &      0.566        & 0.566 &         0.550      &        \textbf{0.680}      \\ \hline
\multicolumn{6}{|c|}{\textbf{\textcolor{blue}{Sydney}}} \\ \hline
16 & \textbf{Accuracy}           &     0.618         &0.639 &     0.561          &      \textbf{0.655}        \\ \hline
17 & \textbf{Precision}          & 0.565             & 0.606 &      0.541         &      \textbf{0.655}        \\ \hline
18 & \textbf{Recall}             &    0.520          & 0.579 &     0.542        &     \textbf{0.664}         \\ \hline
19 & \textbf{F1-Score}         &      \textbf{0.751} &0.740        &        0.638       &       0.690       \\ \hline
20 & \textbf{AUC-ROC}      &      0.618        & 0.638 &         0.540      &        \textbf{0.660}      \\ \hline
\multicolumn{6}{|c|}{\textbf{\textcolor{blue}{Ferguson}}} \\ \hline
21 & \textbf{Accuracy}           &     0.652         & 0.675 &     0.585          &      \textbf{0.705}        \\ \hline
22 & \textbf{Precision}          & 0.598             & 0.634 &      0.549        &      \textbf{0.671}        \\ \hline
23 & \textbf{Recall}             &    0.519          & 0.589 &     0.549         &     \textbf{0.658}         \\ \hline
24 & \textbf{F1-Score}         &     0.782       & 0.778 &        0.676       &       \textbf{0.783}       \\ \hline
25 & \textbf{AUC-ROC}      &      0.652        & \textbf{0.674} &         0.550      &        0.660      \\ \hline
\end{tabular}
\end{table}
\subsection{Results}\label{sec:Res}
In this section, we discuss the micro and macro-analysis of our evaluation for all the incidents using five metrics.

\noindent \textbf{{1. Macro-Analysis}}\label{sec:macro}: Table \ref{Tbl:gw} provides the macro-averaged results of our machine learning models for all the five metrics for \textit{Charlie}, \textit{German}, \textit{Ottawa}, \textit{Sydney}, and \textit{Ferguson} respectively. In general, GCN outperforms other classifiers in all the five metrics. SVM and RF perform better than LSTM in most of the metrics. Considering per incident evaluations, the results of \textit{German} incident outperforms with a significant margin, whereas \textit{Charlie} results exceed with a small margin. In spite of using Stratified K-Fold, the class imbalance problem might have affected the \textit{Charlie} results.

\begin{figure*}[ht!]
\begin{tabular}{|c|c|c|c|c|c|}
\hline
{} & Accuracy & Precision & Recall & F1-Score & AUC-ROC\\\hline
\rotatebox{90}{\hspace{4mm}Charlie}
&
\subfloat{\includegraphics[width=0.35\columnwidth]{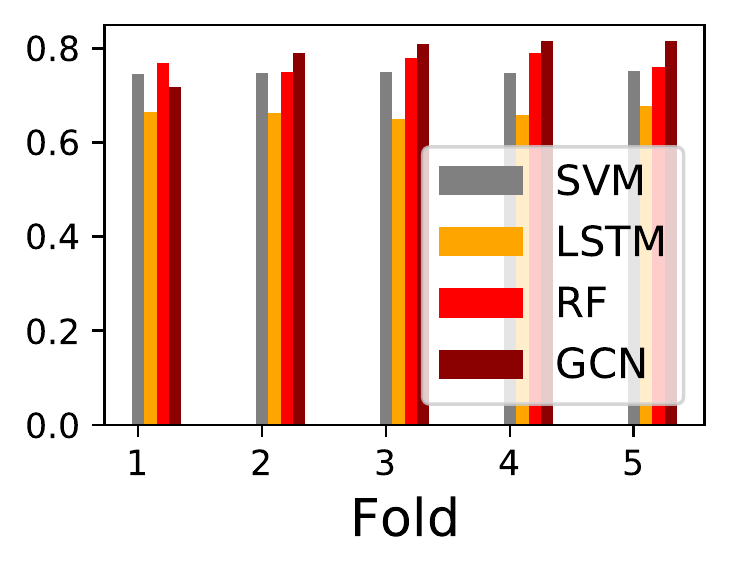}} &
\subfloat{\includegraphics[width=0.35\columnwidth]{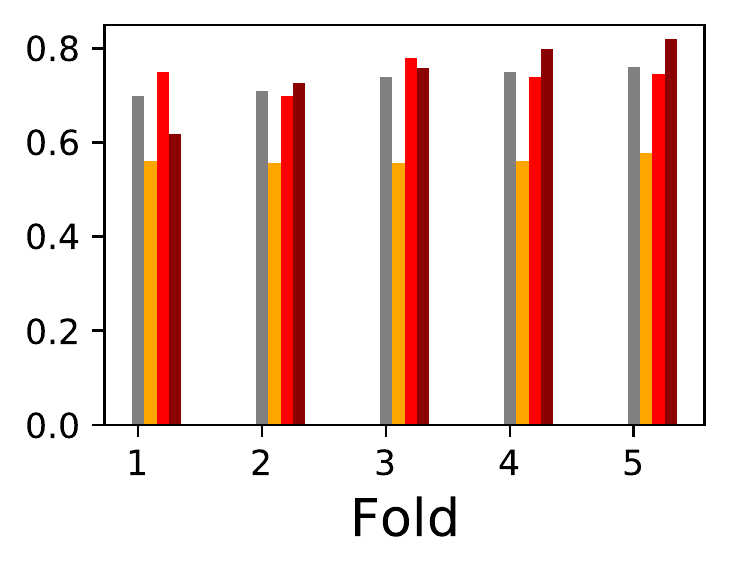}} &
\subfloat{\includegraphics[width=0.35\columnwidth]{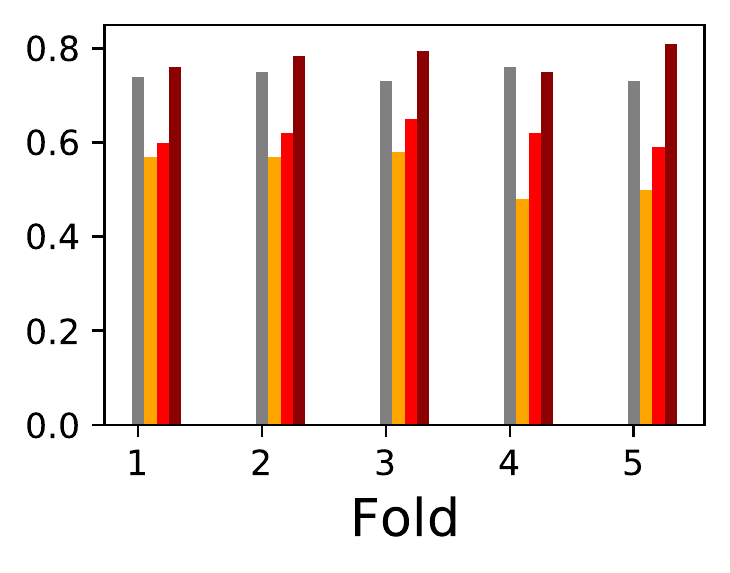}} &
\subfloat{\includegraphics[width=0.35\columnwidth]{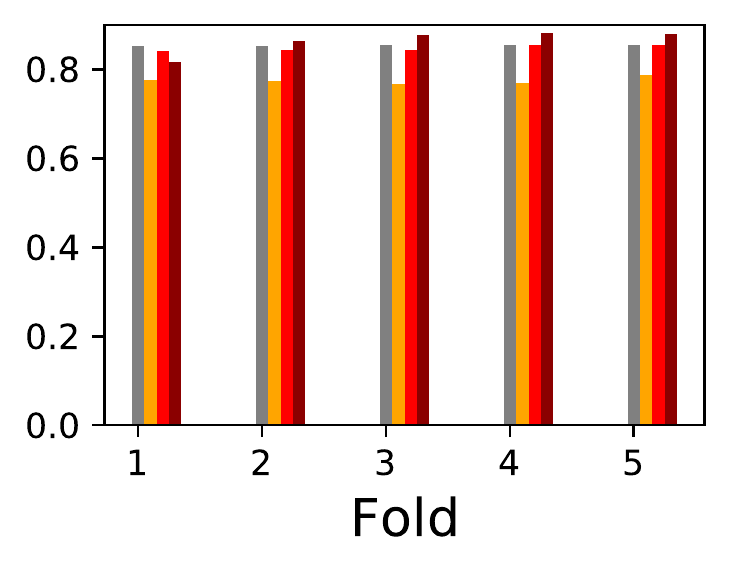}} &
\subfloat{\includegraphics[width=0.35\columnwidth]{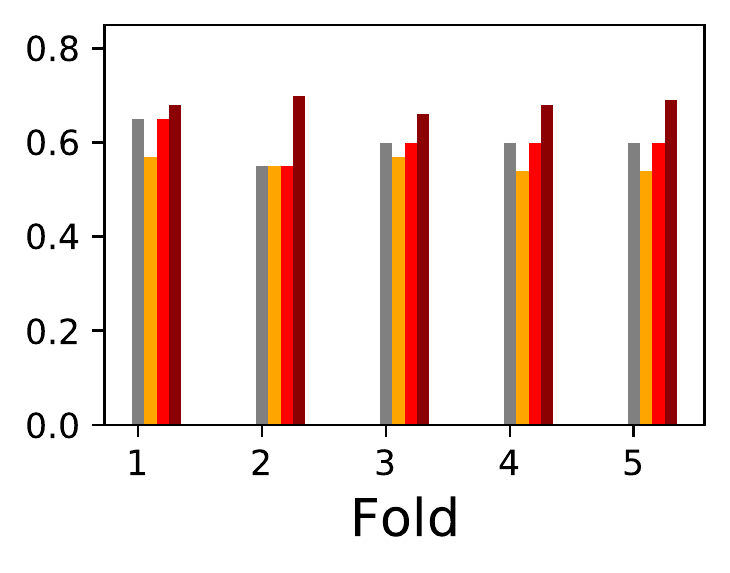}}\\\hline

\rotatebox{90}{\hspace{4mm}German} &
\subfloat{\includegraphics[width=0.35\columnwidth]{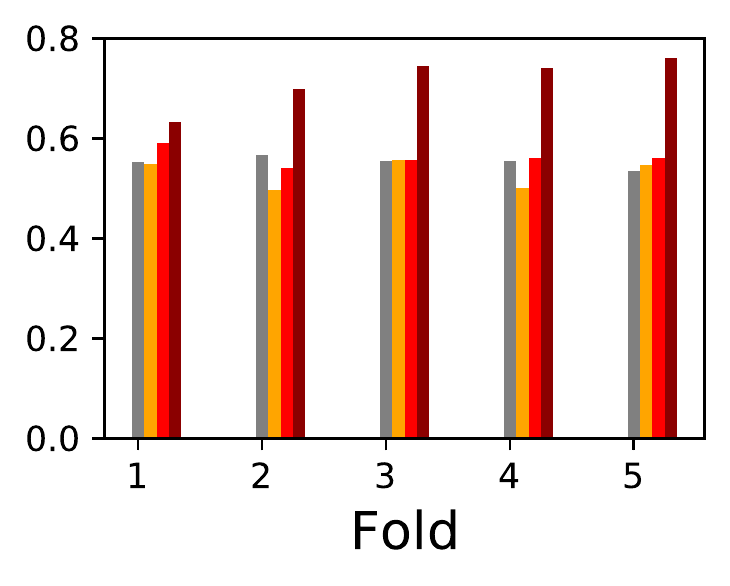}} &
\subfloat{\includegraphics[width=0.35\columnwidth]{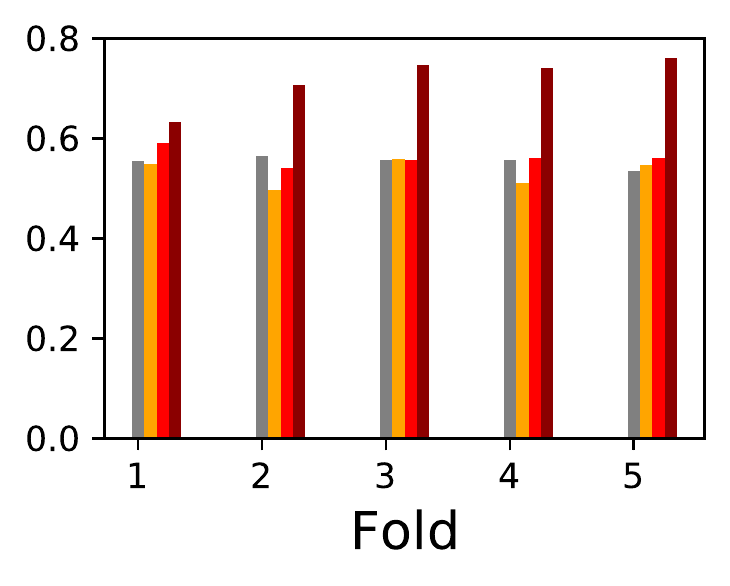}} &
\subfloat{\includegraphics[width=0.35\columnwidth]{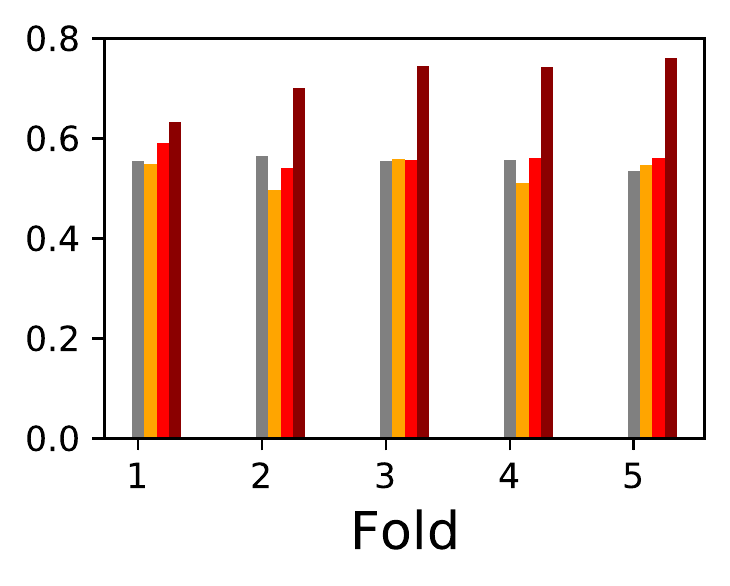}} &
\subfloat{\includegraphics[width=0.35\columnwidth]{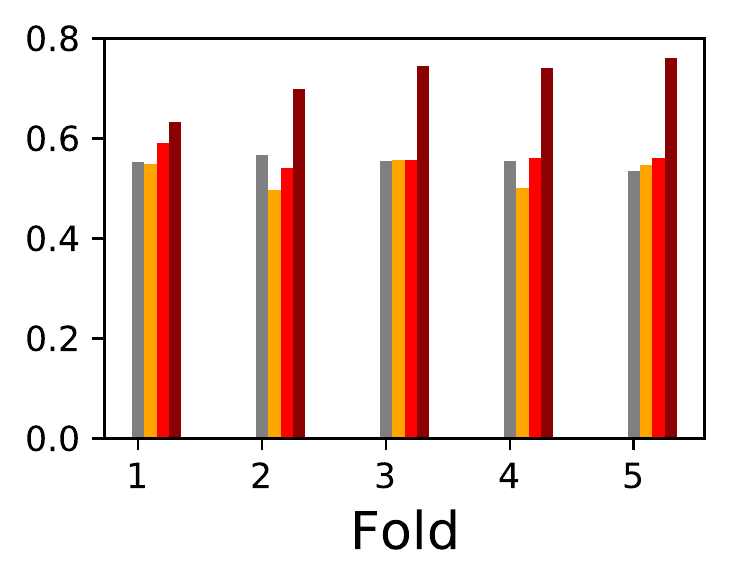}} &
\subfloat{\includegraphics[width=0.35\columnwidth]{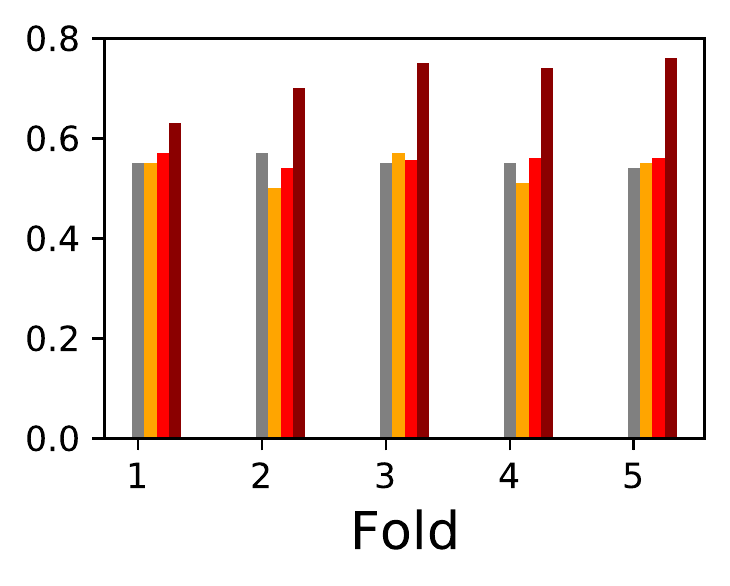}}\\\hline

\rotatebox{90}{\hspace{7mm}Ottawa} &
\subfloat{\includegraphics[width=0.35\columnwidth]{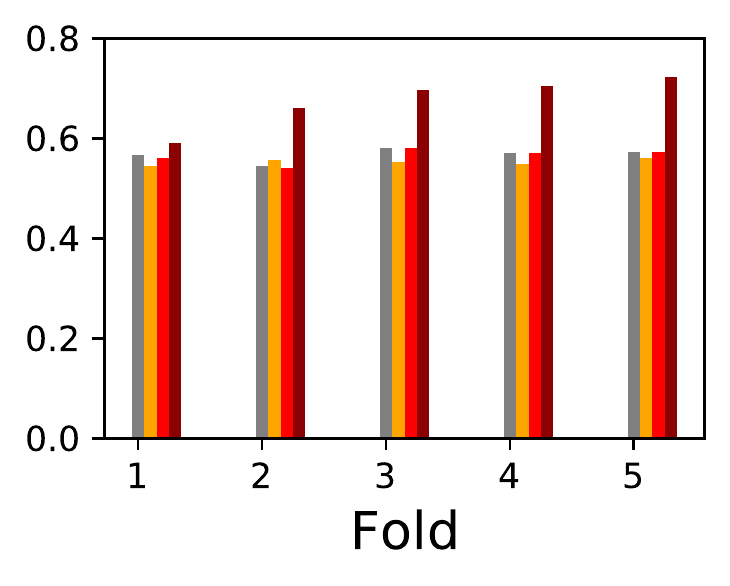}} &
\subfloat{\includegraphics[width=0.35\columnwidth]{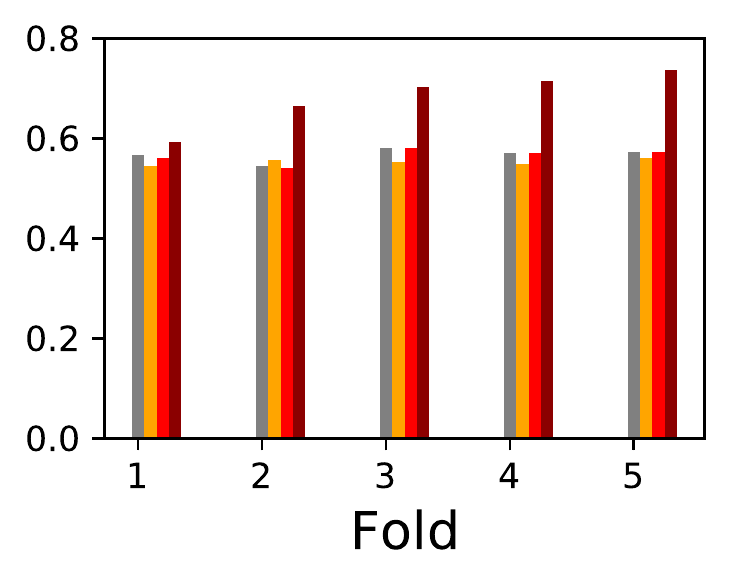}} &
\subfloat{\includegraphics[width=0.35\columnwidth]{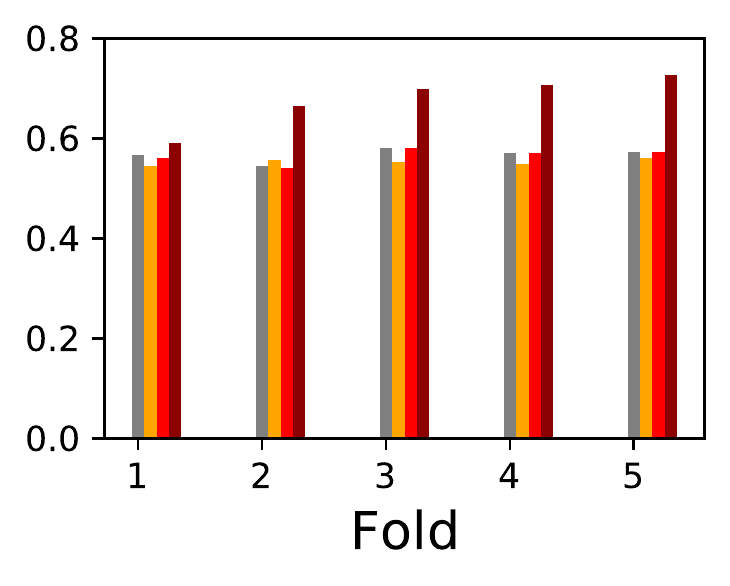}} &
\subfloat{\includegraphics[width=0.35\columnwidth]{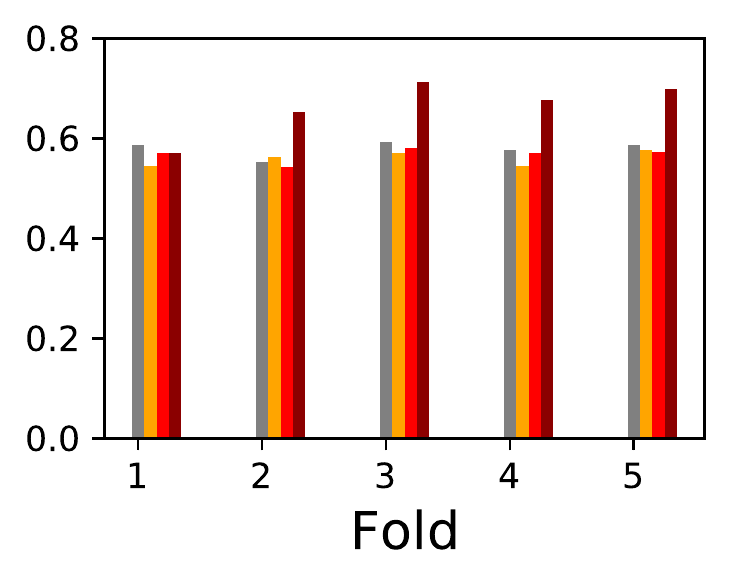}} &
\subfloat{\includegraphics[width=0.35\columnwidth]{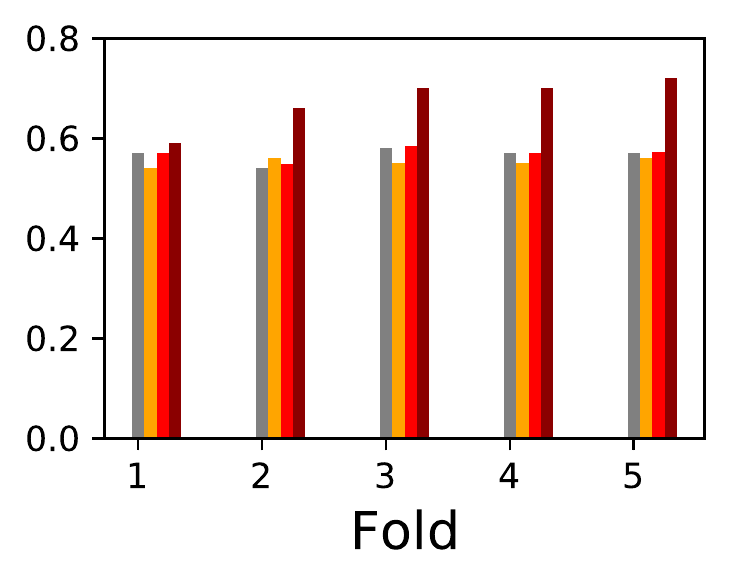}}\\\hline

\rotatebox{90}{\hspace{6mm}Sydney} &
\subfloat{\includegraphics[width=0.35\columnwidth]{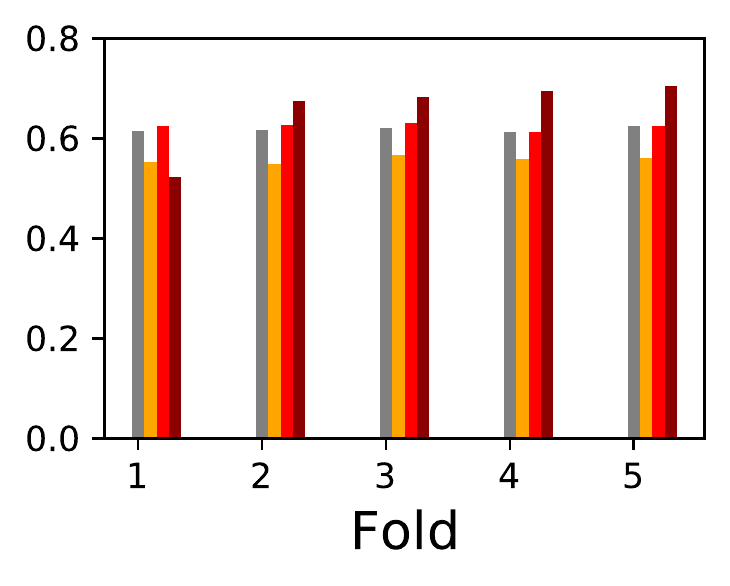}} &
\subfloat{\includegraphics[width=0.35\columnwidth]{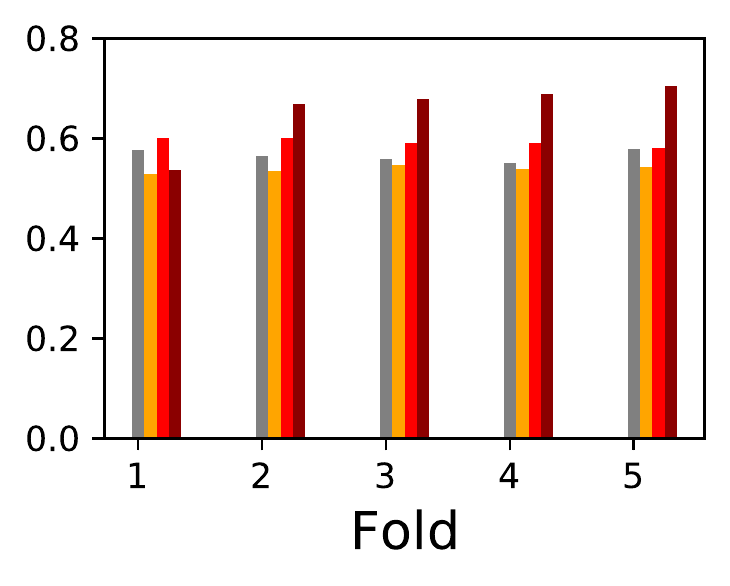}} &
\subfloat{\includegraphics[width=0.35\columnwidth]{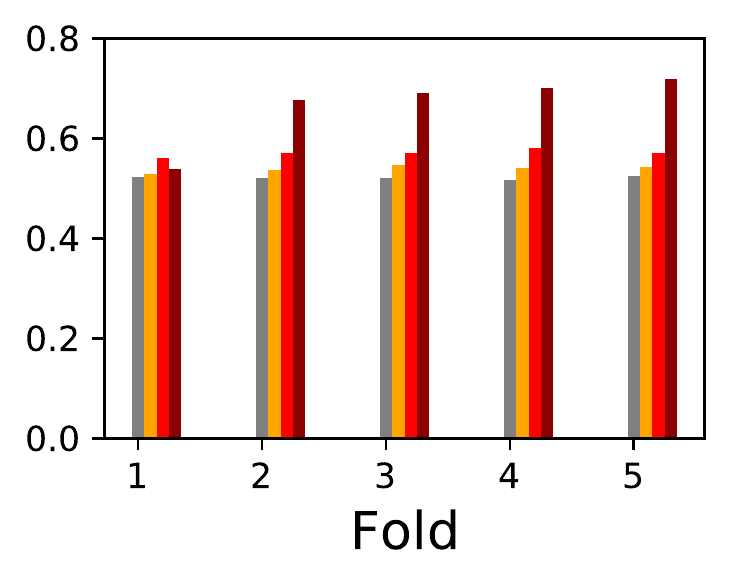}} &
\subfloat{\includegraphics[width=0.35\columnwidth]{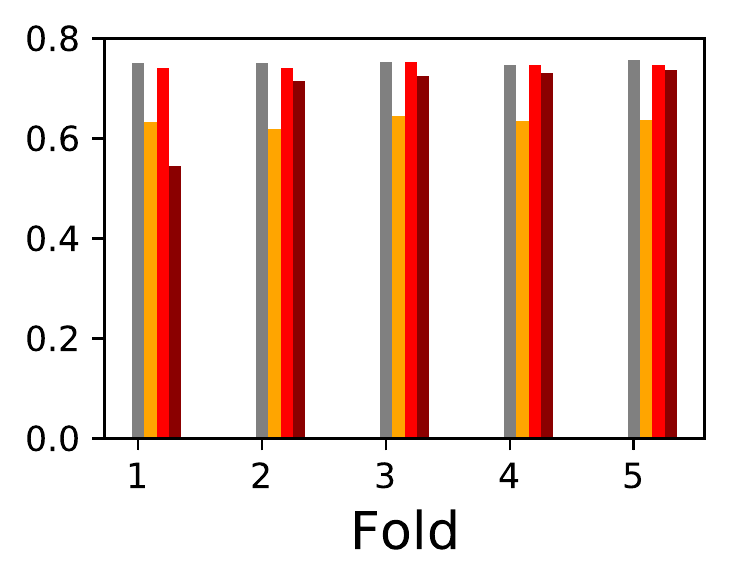}} &
\subfloat{\includegraphics[width=0.35\columnwidth]{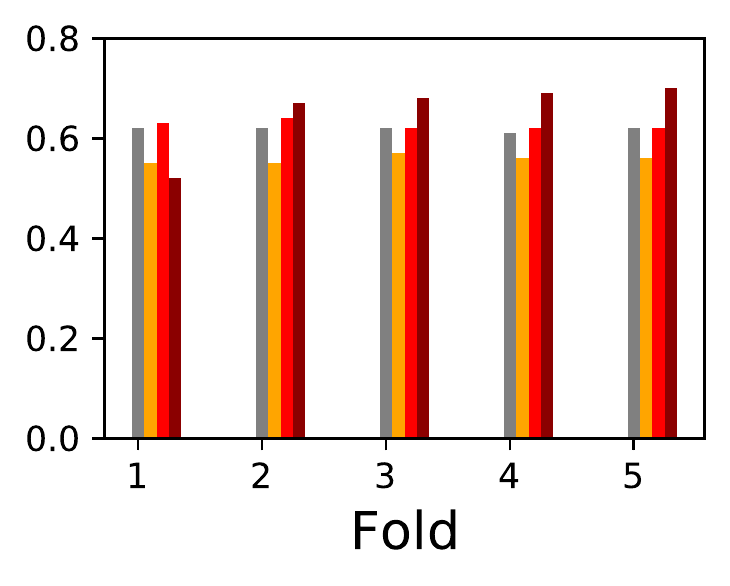}}\\\hline

\rotatebox{90}{\hspace{5mm}Ferguson} &
\subfloat{\includegraphics[width=0.35\columnwidth]{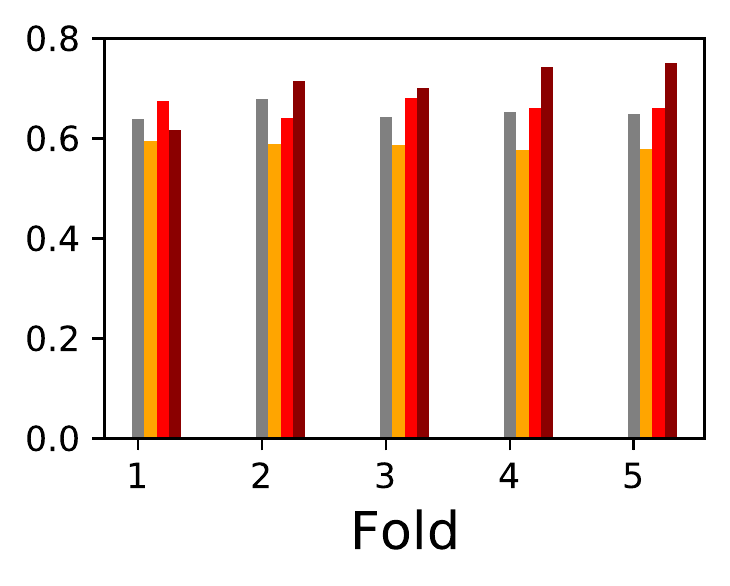}} &
\subfloat{\includegraphics[width=0.35\columnwidth]{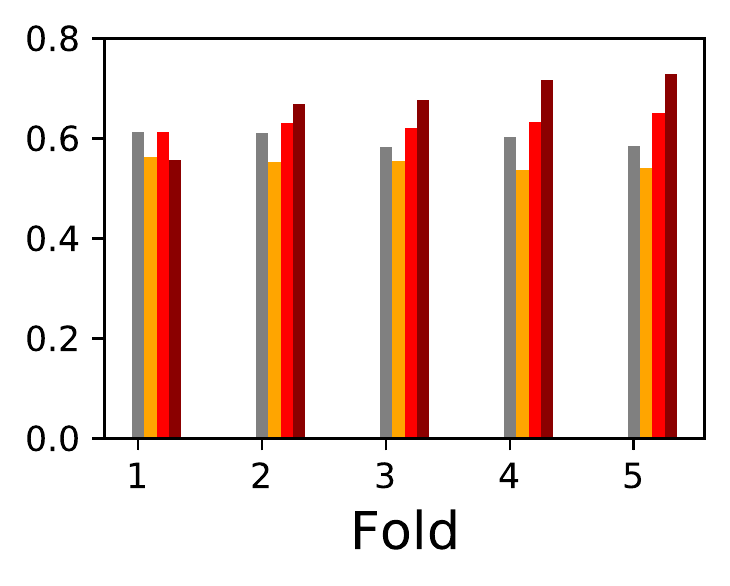}} &
\subfloat{\includegraphics[width=0.35\columnwidth]{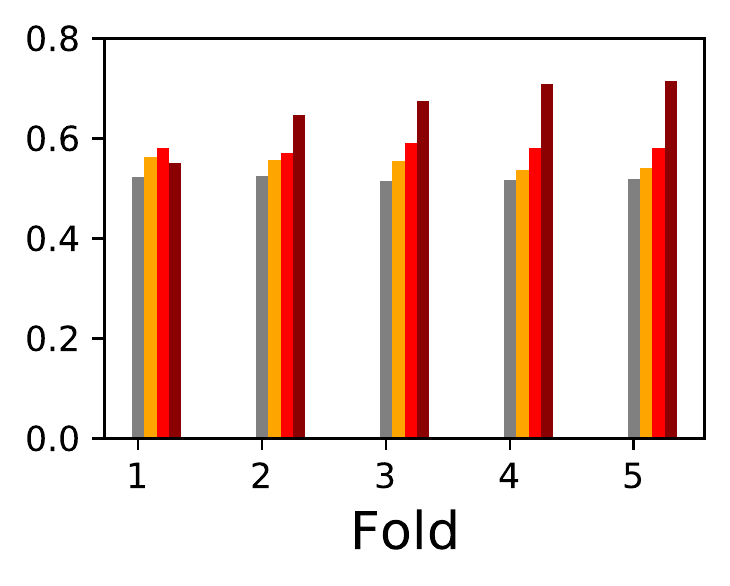}} &
\subfloat{\includegraphics[width=0.35\columnwidth]{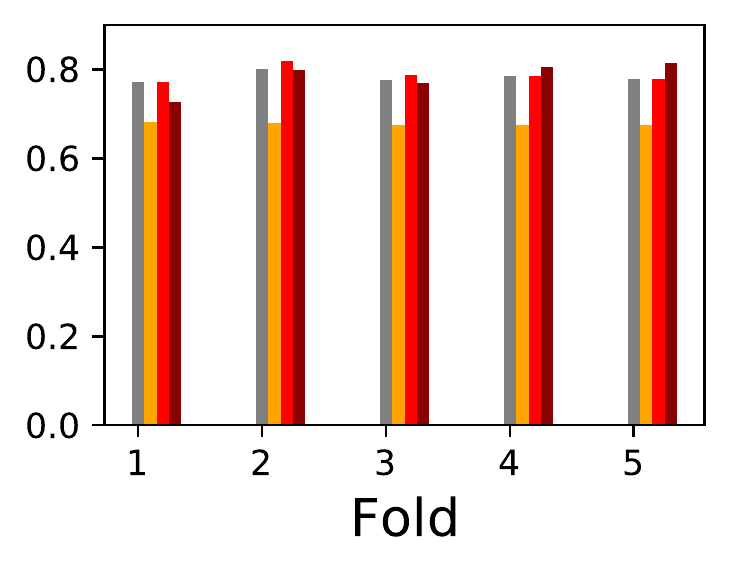}} &
\subfloat{\includegraphics[width=0.35\columnwidth]{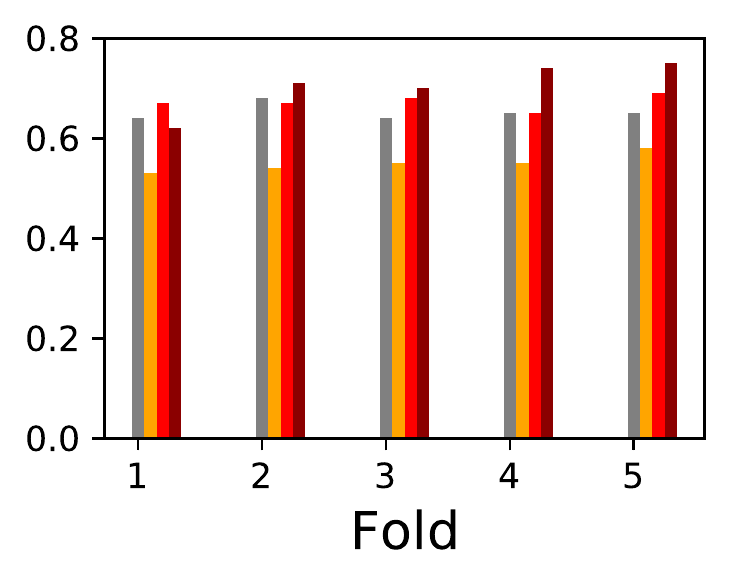}}\\\hline

\end{tabular}
\caption{Performance of GCN model under five-fold cross validation for all the incidents}
\label{fig:cv}
\end{figure*}

\begin{figure*}[!htbp]
     \subfloat[Charlie \label{subfig-1:charlie}]{%
       \includegraphics[width=0.20\textwidth]{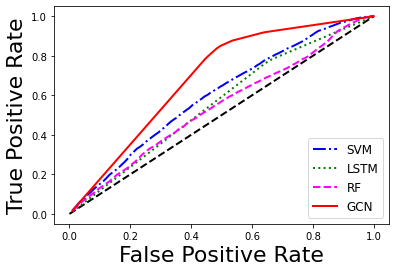}
     }
       \subfloat[German  \label{subfig-1:german}]{%
       \includegraphics[width=0.20\textwidth]{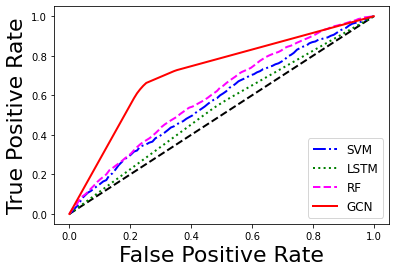}
     }
     \subfloat[Ottawa \label{subfig-2:ottawa}]{%
       \includegraphics[width=0.20\textwidth]{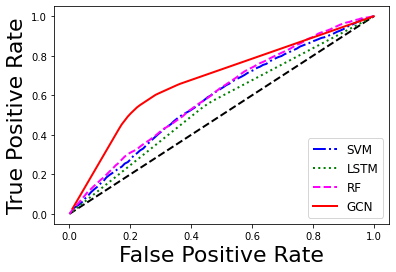}
     }
      \subfloat[Sydney \label{subfig-2:sydney}]{%
       \includegraphics[width=0.20\textwidth]{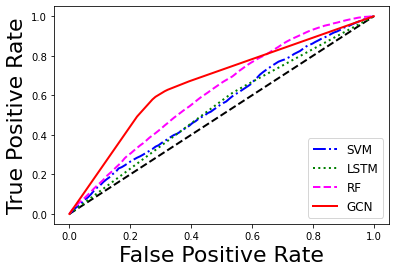}
     }
      \subfloat[Ferguson\label{subfig-2:ferguson}]{%
       \includegraphics[width=0.20\textwidth]{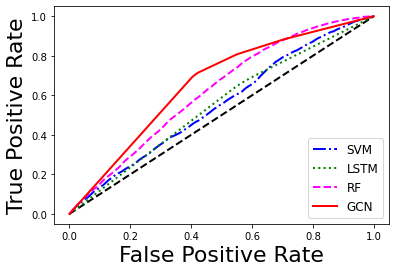}
     }
     \caption{Performance of AUC-ROC Curve on all the five incidents}
     \label{fig:aucroc}
   \end{figure*}

\noindent \textbf{{2. Micro-Analysis}}\label{sec:micro}: Figure \ref{fig:cv} shows the micro-performance of each of the approaches under five metrics for all the incidents in our dataset. Specifically, the x-axis represents the fold number under five-fold cross-validation, whereas the y-axis represents the metrics used for evaluation. To summarize, each incident consisting of five plots depicting Accuracy, Precision, Recall, F1-Score, and AUC-ROC Score for each fold. It can be seen from the plots that, in general, the GCN approach is performing better than the baseline models. However, GCN approach on \textit{German} and \textit{Ottawa} incidents performed significantly better compared to the other two approaches.
This clearly implies that the GCN approach is a natural fit for our problem statement.

Furthermore, we plot the AUC-ROC plot for each of the incidents to understand the micro-performance of the algorithm at different thresholds. Figure \ref{fig:aucroc} shows the AUC-ROC Curve for \textit{Charlie}, \textit{German}, \textit{Ottawa}, \textit{Sydney}, and \textit{Ferguson} respectively. It can be noticed that SVM, RF, and LSTM perform little better than the random model, whereas GCN is better with a good margin. However, RF performs better than the other two baselines in the case of the \textit{Ferguson} incident. Besides, the GCN approach performs well on lower thresholds for \textit{German}, \textit{Ottawa}, and \textit{Sydney}. In contrast, the reverse is the case for \textit{Charlie}, and \textit{Ferguson} indicating the reasons, higher values of Accuracy, Precision, Recall, and F1-Score than AUC-ROC values for the \textit{Charlie} and higher values of Accuracy, Precision, and F1-Score
than AUC-ROC values for the \textit{Ferguson} incident. In all cases, the results are indicative that GCN is able to exploit the relation among the initiator's tweet and its responders, which helped it to perform better.

%% file: conclusion.tex
\section{Conclusion and Future Work}\label{sec:concl}
Identifying possible \textit{rumor spreaders} is crucial as it has been shown that they are the potential sources of \textit{rumor} propagation \cite{volkova2017separating}. In this work, we use the \textbf{PHEME} dataset to identify
possible \textit{rumor spreaders} 
using a weak supervised learning approach. We model this problem as binary classification task by
applying various machine learning models to the transformed rumor spreaders dataset. 
Our results show that  GCN  (compared  to baseline  models)  is  able  to  perform better (to raise red flags for possible \textit{rumor spreaders}) by exploiting relationships of possible \textit{rumor spreaders} who could blend well with \textit{non-rumor spreaders}.
The overall performance of the GCN shows the effectiveness of this approach.
We would like to improve this work through the following multiple plans:
\begin{enumerate}
    \item \textbf{Multiclass problem: }In our present work, we transformed the dataset to study it as a binary classification problem. However, in our extended work, we would like to model this problem as a multiclass prediction to minimize the loss in transformation. 
    \item \textbf{Additional datasets: }
    The transformation of \textit{rumor} dataset into rumor spreaders dataset may induces bias. To overcome that,
    we would like to apply this framework on datasets containing all the tweets posted by a user as opposed to only collecting tweets by a specific topic. 
    \item \textbf{New algorithms: }To extend our research, we plan to apply other graph neural network techniques that works well on unseen graph structures (inductive learning) such as Graph Attention Networks \cite{velivckovic2017graph}, GraphSage \cite{wu2020comprehensive}. 
\end{enumerate}
\section{ACKNOWLEDGMENT}
This  research  is  funded  by H2020 project, SoBigData++, and CHIST-ERA project SAI.